\documentclass[10pt,twocolumn,letterpaper]{article}

\usepackage{iccv}
\usepackage{times}
\usepackage{epsfig}
\usepackage{graphicx}
\usepackage{amsmath}
\usepackage{amssymb}
\usepackage{booktabs}
\usepackage{verbatim}
\usepackage{multirow}
\usepackage{makecell}
\usepackage{gensymb}
\usepackage{enumitem}
\usepackage{cuted}
\usepackage{capt-of}
\usepackage{pifont}
\usepackage[table]{xcolor}
\usepackage{xspace}
\usepackage[accsupp]{axessibility} 

\usepackage[pagebackref=true,breaklinks=true,letterpaper=true,colorlinks,bookmarks=false]{hyperref}

\iccvfinalcopy 

\def\iccvPaperID{1881} 

\ificcvfinal\fi

\author{
Noah Stier$^{1,2}$ \hspace{3pt}
Anurag Ranjan$^{1}$ \hspace{3pt}
Alex Colburn$^{1}$ \hspace{3pt}
Yajie Yan$^{1}$ \hspace{3pt} \\
Liang Yang$^{1}$ \hspace{3pt}
Fangchang Ma$^{1}$ \hspace{3pt}
Baptiste Angles$^{1}$ \hspace{3pt}\\[1em]
$^1$Apple \hspace{6pt}
$^2$University of California, Santa Barbara \hspace{3pt}
}

\usepackage{overpic}
\usepackage{multirow}
\usepackage{dirtree}
\newcommand{\OURS}{\textit{FineRecon}\xspace}

\newcommand{\rone}[1]{\textcolor[rgb]{1, 0, 0}{\textbf{R1}}}
\newcommand{\rtwo}[1]{\textcolor[rgb]{0, 1, 0}{\textbf{R2}}}
\newcommand{\rthree}[1]{\textcolor[rgb]{0, 0, 1}{\textbf{R3}}}

\newcommand{\emptysquare}{$\square$}
\newcommand{\checkedsquare}{\makebox[0pt][l]{\raisebox{1pt}[0pt][0pt]{\normalsize\hspace{1pt}\cmark}}$\square$}
\newcommand{\cmark}{\ding{51}}%

\newcommand{\firstcell}{\cellcolor{red!35}}
\newcommand{\secondcell}{\cellcolor{orange!35}}

\begin{document}

\title{FineRecon: Depth-aware Feed-forward Network for Detailed 3D Reconstruction}

\maketitle
\ificcvfinal\thispagestyle{empty}\fi

\maketitle

\begin{abstract}

Recent works on 3D reconstruction from posed images~\cite{atlas, vortx, neuralrecon} have demonstrated that direct inference of scene-level 3D geometry without test-time optimization is feasible using deep neural networks, showing remarkable promise and high efficiency.  
However, the reconstructed geometry, typically represented as a 3D truncated signed distance function (TSDF), is often coarse without fine geometric details.
To address this problem, we
propose three effective solutions for improving the fidelity of inference-based 3D reconstructions.
We first present a resolution-agnostic TSDF supervision strategy to provide the network with a more accurate learning signal during training, avoiding the pitfalls of TSDF interpolation seen in previous work. 
We then introduce a depth guidance strategy using multi-view depth estimates to enhance the scene representation and recover more accurate surfaces.
Finally, we develop a novel architecture for the final layers of the network, conditioning the output TSDF prediction on high-resolution image features in addition to coarse voxel features, enabling sharper reconstruction of fine details. Our method, \OURS \footnote{\href{https://github.com/apple/ml-finerecon}{https://github.com/apple/ml-finerecon}},
produces smooth and highly accurate reconstructions, showing significant improvements across multiple depth and 3D reconstruction metrics.

\end{abstract}
\section{Introduction}
\label{sec:intro}

Reconstruction of 3D scenes from posed images is a long-standing problem in computer vision, with many applications such as autonomous driving, robotic navigation, and digital 3D asset creation. The traditional approach is to estimate depth maps over the input images using multi-view stereo (MVS), and then fuse them together to form a unified 3D model~\cite{furukawa2015multi, galliani2015massively, colmap}. However, the fusion process commonly results in missing geometry or artifacts in areas where the depth maps do not agree, due to effects such as occlusion, specularity, and transparent or low-texture surfaces. Recently, an alternative method has been proposed to address this issue in Atlas~\cite{atlas}, which back-projects learned image features onto a voxel grid and directly predicts the scene's truncated signed distance function (TSDF) using a 3D convolutional neural network (CNN).
The main advantage is that the CNN can learn to produce smooth, consistent surfaces, and to fill in holes that would otherwise result from low-texture regions and occlusion.
Several methods have proposed improvements to this framework \cite{transformerfusion, choe2021volumefusion, vortx, neuralrecon}, consistently pushing the state of the art in reconstruction accuracy. However, despite these efforts, the reconstructions produced by these methods remain coarse. We identify three key factors restricting the accuracy and level of detail in prior works, and we introduce solutions to address them, demonstrating their effectiveness within a new system: \emph{\OURS}.

\begin{figure}[t!]
\begin{center}
    \begin{overpic}
    [width=\linewidth]
    {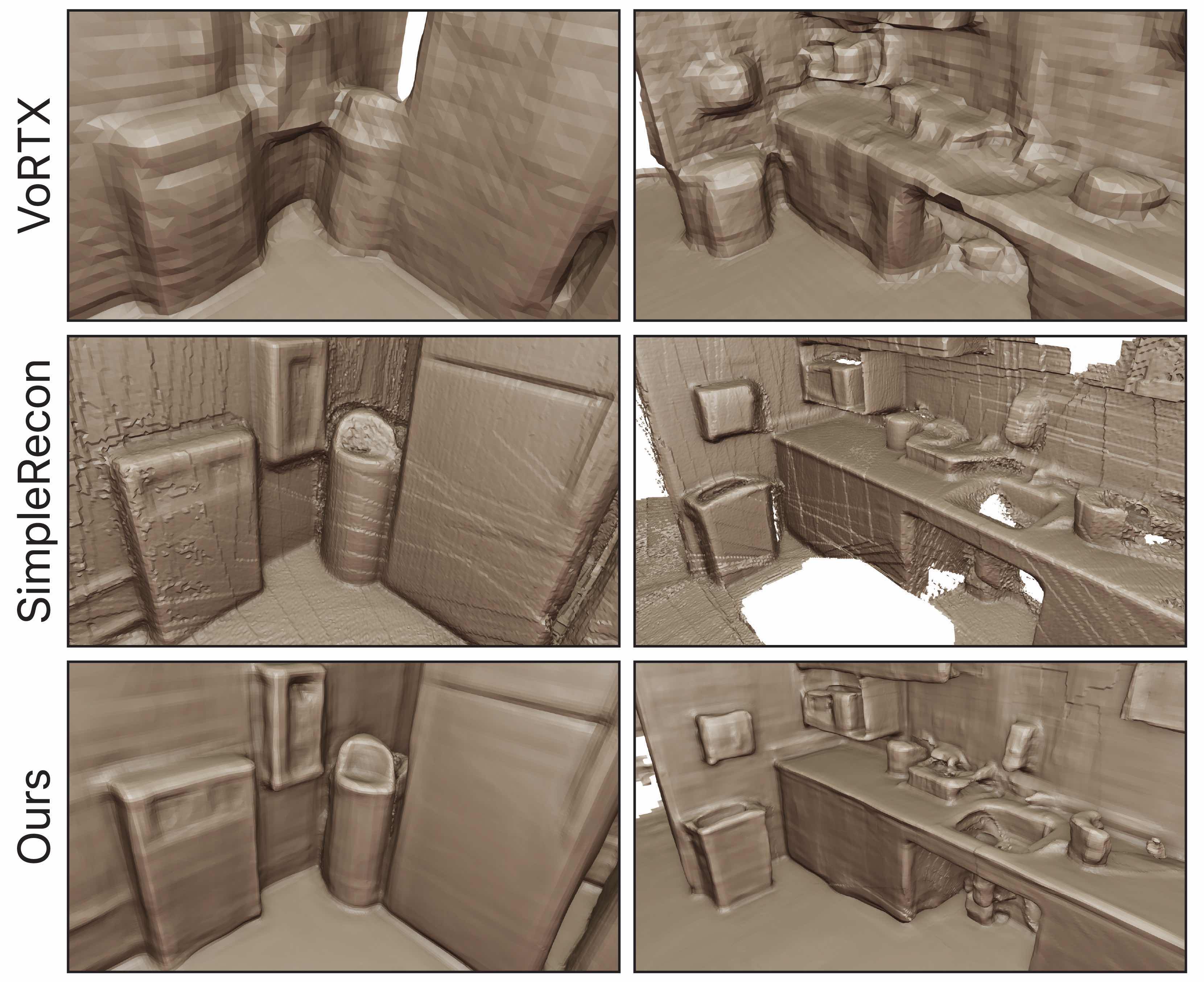}
    \end{overpic}
    \end{center}
    \vspace{-1em}
    \captionof{figure}{
        Our method, \textbf{FineRecon}, recovers highly detailed and coherent geometry relative to state-of-the-art methods. Our contributions of lossless ground truth sampling, depth-aware feature volume, and point backprojection result in smooth surfaces that preserve high-frequency structures without creating strong artifacts.
    }
    \vspace{-1em}
\label{fig:teaser}
\end{figure}

First, existing works use tri-linear interpolation to resample the ground-truth TSDF to align with the model's voxel grid during training~\cite{atlas, vortx, neuralrecon}. This allows supervision of the model's TSDF predictions at each voxel center, even when the voxel centers do not coincide with the pre-computed ground-truth TSDF points. However, resampling via tri-linear interpolation corrupts detail in the training data, because distance fields are not linear when non-planar geometry such a corner is present, as shown in Fig.~\ref{fig:tsdf}.
We avoid this issue by making supervised predictions only at the exact points where the ground-truth TSDF is known. This supervision change comes at no extra cost, and it results in greatly improved visual detail as well as a relative reduction in average chamfer distance between reconstruction and ground truth of over 10\%. 

Second, prior work~\cite{transformerfusion, choe2021volumefusion, atlas, vortx, neuralrecon} uses dense back-projection, sampling a feature from each input image in each voxel. This causes blurring in the back-projection volume, which increases the difficulty of extracting accurate surface locations. To address this, our method uses an initial multi-view stereo depth estimation step, after which the depth estimates are used to enhance the feature volume and guide the 3D CNN toward areas of high surface likelihood. We show that this step significantly increases the quality of the reconstructions produced by our system.

Third, because of the high computational cost of 3D CNNs, it is expensive to increase the voxel resolution. Existing works use voxel sizes of 4cm or larger~\cite{transformerfusion, choe2021volumefusion, atlas, vortx, neuralrecon}, which is not enough to resolve the level of geometric detail visible in natural images at ranges of a few meters. To remedy this, we propose a new method to query the TSDF prediction at any point in $\mathbb{R}^3$, conditioned on the CNN grid features \textit{and} image features projected directly to the query point. This reduces aliasing and allows our model to resolve sub-voxel detail. Furthermore, this enables reconstruction at arbitrary resolution without re-training.

\OURS achieves state-of-the-art performance on the challenging ScanNet dataset, as measured by 3D mesh metrics and rendered 2D depth metrics. We further show that it produces substantially improved visual detail with reduced artifacts relative to prior work.

\textbf{Contributions.}
The main contributions of this paper are:
\vspace{-1.5em}
\begin{itemize}
  \setlength\itemsep{0.05em}
    \item We increase the accuracy of the training data using resolution-agnostic TSDF supervision, allowing \OURS to reconstruct details with higher fidelity.
    \item We improve reconstruction accuracy using a novel MVS depth-guidance strategy, augmenting the back-projection volume with an estimated TSDF fusion channel.
    \item We enable the reconstruction of sub-voxel detail with a novel TSDF prediction architecture that can be queried at any 3D point, using \textit{point back-projected} fine-grained image features.
\end{itemize}

\section{Related Work}
\label{sec:related}

\paragraph{Multi-view stereo.}
3D reconstruction is traditionally posed as per-pixel depth estimation \cite{furukawa2015multi, galliani2015massively, colmap}. While recent works have shown strong results \cite{im2019dpsnet, simplerecon}, a known drawback is that the estimation of each depth map is independent, so continuity across frustum boundaries is not enforced, and this often leads to artifacts. Solutions have been proposed \cite{deepvideomvs, hou2019multi, rich20213dvnet}, but it is still an open problem.

\textbf{Feed-forward 3D reconstruction.}
An effective recent strategy is to perform volumetric reconstruction directly in scene space using feed-forward neural networks \cite{transformerfusion, choe2021volumefusion, atlas, vortx, neuralrecon}. In this line of research, image features are encoded by a 2D CNN and densely back-projected into a global feature volume, then a 3D CNN predicts the scene TSDF. These models can generalize to new scenes at inference time without computationally-demanding test-time optimization, and they can produce smooth and complete reconstructions. However, they tend to blur out surface details and omit thin structures. In contrast, ours is the first of these volumetric TSDF methods to reconstruct accurate sub-voxel detail. 3D-Former~\cite{yuan2023monocular}, a concurrent work, replaces the 3D CNN with a transformer, an orthogonal direction that could potentially be combined with our method.




\textbf{Geometric priors in neural radiance fields.}
Recent novel-view synthesis methods based on neural radiance fields~\cite{nerf} have shown remarkable 2D rendering quality, typically relying on time-consuming per-scene optimization to obtain good results. These methods have also been extended to predict surface geometry \cite{wang2021neus, yariv2021volume}, and it has been shown that geometric priors, such as depth and normal maps, can boost performance. For instance, multiple works~\cite{efficient, denseprior, nerfingmvs} use depth estimates to improve ray sampling efficiency. MonoSDF~\cite{yu2022monosdf} uses monocular depth and normal estimation as pseudo-ground truth to supervise its predicted SDF. Similarly, our method uses guidance from a depth-prediction network, but with a focus on interactive reconstruction speed and generalization to new scenes.

\textbf{Geometric priors in feed-forward networks.}
Only a few previous methods based on feed-forward networks have incorporated geometric priors. MonoNeuralFusion \cite{mononeuralfusion} uses monocular normal estimates but does not explore depth priors. Similar to our approach, VolumeFusion \cite{choe2021volumefusion} uses depth as an additional feature by fusing it into a density volume and concatenating it with the image feature volume. However, density does not encode the difference between occluded space and observed free space, and it omits crucial information about inward vs. outward surface orientation. In contrast, we explore multiple forms of depth guidance, finding that guidance by TSDF fusion outperforms the density-based variant and others (Table \ref{tab:ablation_depth}). CVRecon \cite{feng2023cvrecon}, a concurrent work, uses frustum-aligned cost volumes in addition to global features, which is related to our work but does not explicitly predict depth.

\begin{figure*}[t]
\begin{center}
    \begin{overpic}
    [width=\linewidth]
    {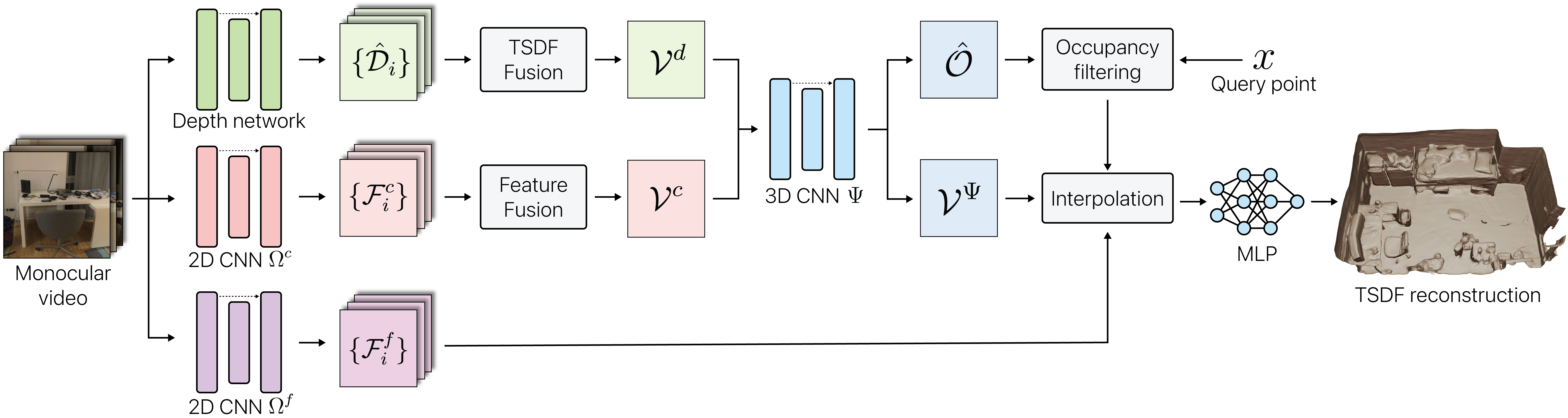}
    \end{overpic}
    \end{center}
    \vspace{-1em}
    \caption{
        \textbf{Model overview.} Given a monocular RGB image sequence, we use a pre-trained depth network to estimate depth images $\mathcal{\hat{D}}$. In the meantime we also extract image features $\mathcal{F}^c$ for global volume fusion and $\mathcal{F}^f$ for point back-projection. The depth $\mathcal{\hat{D}}$ is then fused into an initial, approximate TSDF volume $\mathcal{V}^d$, and $\mathcal{F}^c$ is back-projected into a feature volume $\mathcal{V}^c$ using the camera parameters. The two volumes are then concatenated and fed to a 3D CNN $\Psi$ to produce the global feature volume $\mathcal{V}^\Psi$ and the coarse occupancy grid $\hat{\mathcal{O}}$ (used to accelerate inference). Finally, at any query point $x \in \mathbb{R}^3$, we sample high-resolution image features $\mathcal{F}^f$ and concatenate with the interpolated voxel feature $\mathcal{V}^\Psi(x)$. This is passed to an MLP to predict the final TSDF value.
    }
\label{fig:architecture}
\end{figure*}

\section{Method}
\label{sec:method}

Given a set of input images $\{I_i\}$ along with their camera poses $\{P_i\}$ and intrinsics $\{K_i\}$, we seek to compute an estimate $\hat{S}$ of the true scene TSDF $S$. We train a deep-learning model $\Phi$ to perform this mapping, $\hat{S} = \Phi(\{I_i\}, \{P_i\}, \{K_i\})$.

\begin{figure}[t]
\begin{center}
    \begin{overpic}
    [width=0.94\linewidth]
    {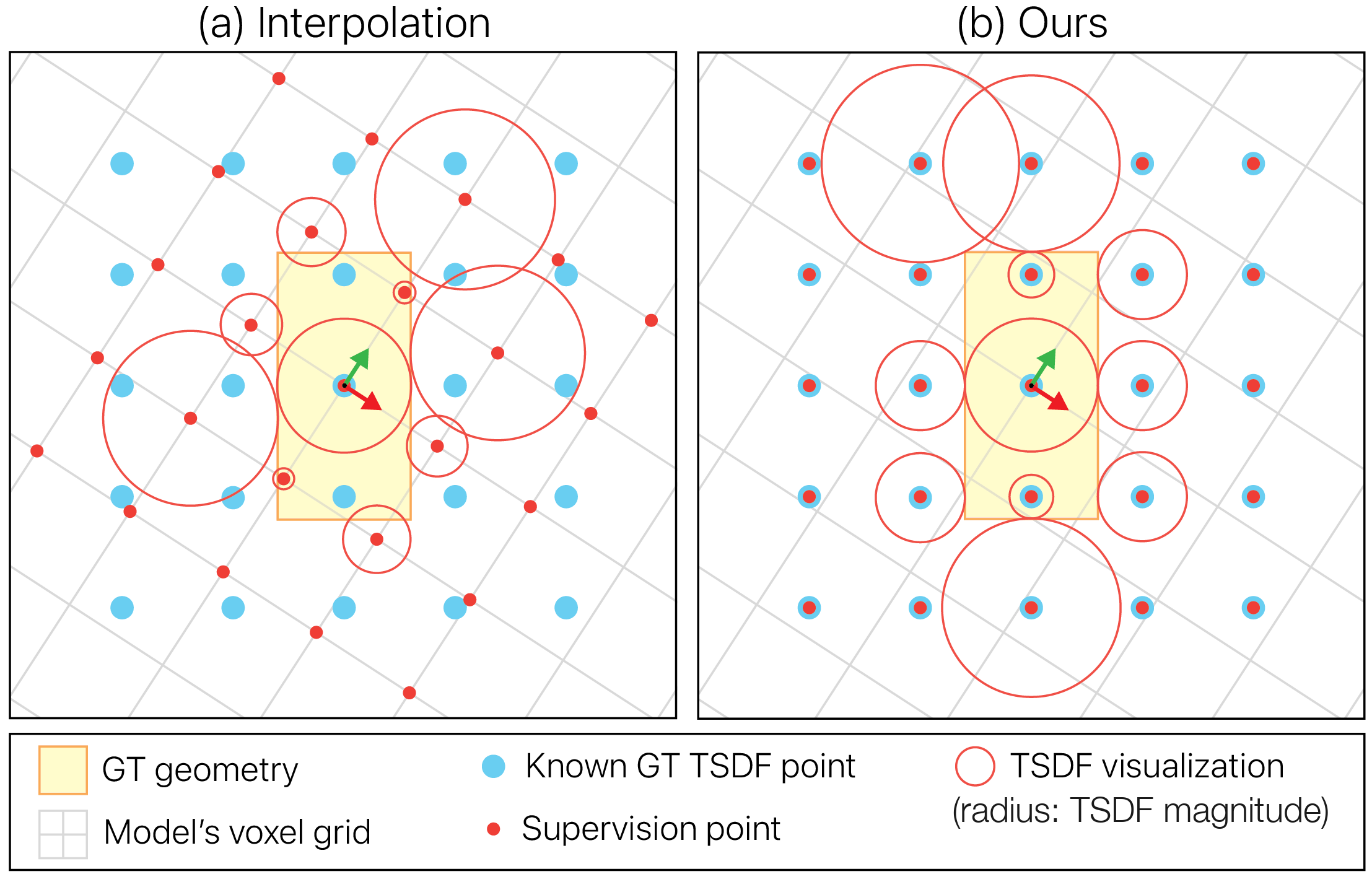}
    \end{overpic}
    \end{center}
    \vspace{-1em}
    \caption{
        \textbf{Illustration of our improved, resolution-agnostic TSDF supervision.}
        With rotational augmentation during training, the model's voxel grid (gray) does not align with the points (blue) where the ground-truth TSDF is known. Previous work uses linear interpolation (a) to estimate the TSDF at the voxel centers for supervision, but this leads to errors: the ground-truth geometry (yellow) is intersected by the interpolated TSDF values (red circles). In contrast, our model can be supervised at any 3D point. Therefore, regardless of the feature grid orientation, we can supervise the model at the exact points where the ground-truth is known (b). This reduces noise during training and increases reconstruction accuracy (see Table~\ref{fig:ablation_main}).
        }
\vspace{-1em}
\label{fig:tsdf}
\end{figure}

\subsection{Model overview}

Our model (illustrated in Fig. \ref{fig:architecture}) extracts a set of 2D features $\{\mathcal{F}_i^c\}$ from each input image using a convolutional neural network (CNN), $\Omega^c$. We then use the camera parameters to back-project $\{\mathcal{F}_i^c\}$ onto a 3D voxel grid. The back-projection process is augmented using depth estimates $\{\hat{\mathcal{D}}_i\}$ to produce the depth-guided feature volume $\mathcal{V}^g$, as described in Section~\ref{subsec:depth_guidance}. We process $\mathcal{V}^g$ using a 3D CNN $\Psi$ to produce a new feature volume $\mathcal{V}^\Psi$, which we can query for $\hat{S}(q)$ at any 3D point $q \in \mathbb{R}^3$ using the interpolation strategy defined in Section~\ref{subsec:highres_inference}. To extract a surface mesh, we query the reconstruction on a grid of arbitrary resolution to produce a TSDF volume, from which a mesh can be extracted with Marching cubes~\cite{marchingcubes}.

\subsection{Key improvements}

\subsubsection{Resolution-agnostic TSDF supervision}
\label{subsec:improved_supervision}

\begin{figure*}[t]
\begin{center}
    \begin{overpic}
    [width=\linewidth]
    {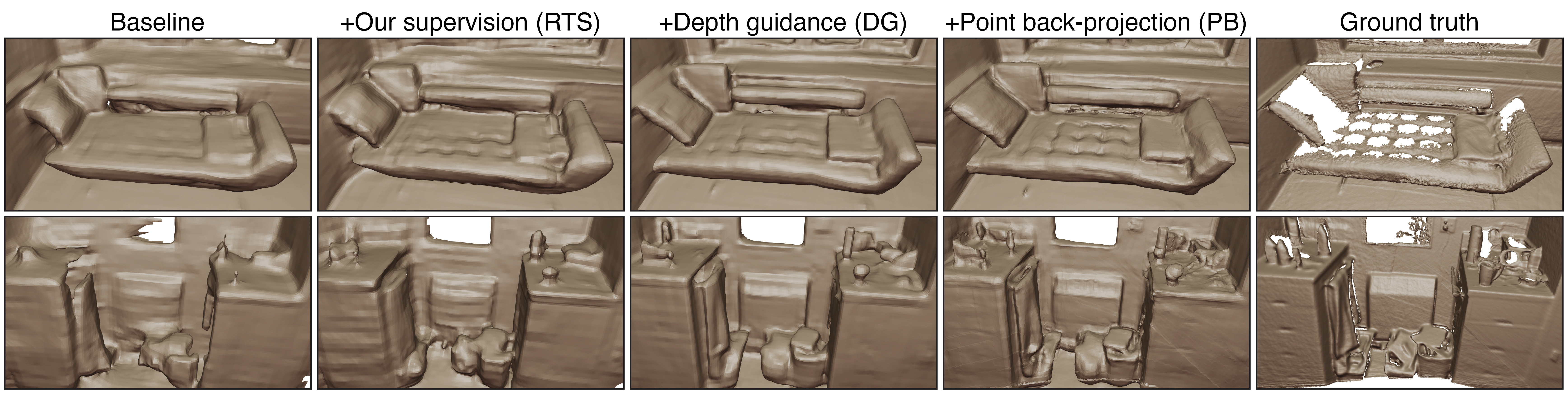}
    \end{overpic}
    \end{center}
    \vspace{-1em}
    \caption{
        \textbf{Ablation of our contributions.} Compared to the baseline result (same as Table \ref{tab:ablation_main}, row (i)), our improved supervision signal provides an overall increase in detail and accuracy at negligible added cost. Inconsistent areas (e.g. right side of couch, towels) are further refined by adding depth guidance. Finally, point back-projection provides higher effective resolution for capturing high-frequency and sub-voxel structures.
}
    \vspace{-0.5em}
\label{fig:ablation_main}
\end{figure*}

During training, as in previous works \cite{atlas, vortx, neuralrecon} we randomly orient the coordinate system of the feature volume relative to $S$ using $\pm 3^\circ$ rotations about the horizontal axis and $\pm 180^\circ$ rotations about the gravitational axis. Thus, the voxel centers may not coincide with the points $\{x\}$ where $S$ is known. This is typically addressed using linear interpolation on $S$ to estimate the ground-truth TSDF at the voxel centers. However, this process is not accurate, since with non-planar geometry the TSDF is a non-linear function of space. Interpolation therefore corrupts the ground-truth TSDF, as shown in Fig. \ref{fig:tsdf}. In theory this effect can be minimized by sampling the ground truth on a very high-resolution grid, but this greatly increases training cost. To preserve the accuracy of the ground truth with no added cost, we instead supervise only at the points $\{x\}$ where the ground truth is known, so that no interpolation is required. This decouples the accuracy of our ground truth from its sampling rate, rendering it \textbf{resolution-agnostic}. To support this, our model must be able to estimate the TSDF at any point in $\mathbb{R}^3$, which we achieve using the strategy outlined in Section \ref{subsec:highres_inference}. This simple supervision change enables our model to reconstruct significantly more visual detail (see Fig. \ref{fig:ablation_main}).

\vspace{-0.4cm}
\subsubsection{Depth guidance}
\label{subsec:depth_guidance}
In order to localize image features in 3D space, we sample a pixel feature at each voxel from each available input image: 
\begin{equation}
    \tilde{\mathcal{V}}^c_i(q) = \alpha(\mathcal{F}_i^c, K_iP_iq_h),
\end{equation}
where $\alpha(F, u)$ represents sampling a feature from 2D map $F$ at pixel location $u$, and $q_h$ is the voxel center in homogeneous coordinates. We then reduce using a per-channel mean across views to form one feature vector per-voxel: 
\begin{equation}
    \mathcal{V}^c(q) = \frac{1}{|\{I_i\}|} \sum_{i} \tilde{\mathcal{V}}^c_i(q)
\end{equation}
As an additional signal in recovering the scene surfaces, we propose to inject a multi-view depth prior using depth estimates from an MVS system $M$. As MVS is a well-studied problem, we treat $M$ as an off-the-shelf component, and in the Supplementary we study the sensitivity of our system to depth noise and choice of $M$. We fuse $\hat{\mathcal{D}}$ into scene space using the standard TSDF fusion \cite{curless1996volumetric} to form $\mathcal{V}^d$.
We then concatenate $\mathcal{V}^d$ as an extra channel in the back-projection volume. Our depth-guided back-projection volume is thus defined as
\begin{equation}
    \mathcal{V}^g = [\mathcal{V}^c, \mathcal{V}^d].
\end{equation}
In our experiments (see Table \ref{tab:ablation_depth}), the depth-guided feature volume $\mathcal{V}^g$ shows significantly improved results with respect to image features alone ($\mathcal{V}^c$), or depth inputs alone ($\mathcal{V}^d$). With naive application of the depth guidance, we find that this additional signal increases our network's propensity to over-fit to the training data, relying too heavily on the depth guidance which is often inaccurate. To address this, we scale each predicted depth map by a factor sampled uniformly in the range $[0.9, 1.1]$ as a data augmentation during training. This reduces over-fitting and encourages the network to learn to use the image features in regions where the depth maps are unreliable.

We additionally experiment with several strategies for using the depth estimate to directly modulate the image feature back-projection. However, as shown in our experiments, our TSDF fusion approach outperforms these methods by a large margin (see Table \ref{tab:ablation_depth}).

\vspace{-0.4cm}
\subsubsection{Point back-projection TSDF inference}
\label{subsec:highres_inference}

We use tri-linear interpolation to sample the 3D CNN's output feature volume $\mathcal{V}^\Psi$ at any query point $q$. This results in a continuous-valued feature $\tilde{\mathcal{V}}^\Psi = \Lambda(\mathcal{V}^\Psi, q)$ where $\Lambda$ represents tri-linear interpolation. Directly estimating $\hat{S}$ from this feature is severely limited in its ability to reconstruct sub-voxel detail, since the effective resolution of $\tilde{\mathcal{V}}^\Psi$ is still constrained to the size of the voxels.

We improve on this paradigm using an additional \textbf{point back-projection} step to directly sample image features $\mathcal{W}(q)$ at the point $q$. This step is identical to the depth-guided back-projection outlined in Section \ref{subsec:depth_guidance}, \textit{except} that the point $q$ is no longer constrained to be a voxel center. The effective resolution of this new continuous-valued feature is thus determined by the resolution of the 2D image feature grid rather than the 3D voxel size. Assuming high-enough 2D feature resolution, $\mathcal{W}(q)$ thus carries much finer-grained information than the linearly-interpolated voxel feature $\tilde{\mathcal{V}}^\Psi(q)$, complementing the 3D CNN's ability to produce smooth and context-informed features. We concatenate $\mathcal{W}$ with $\tilde{\mathcal{V}}^\Psi$ as the input to a multi-layer perceptron (MLP) $\theta_S$:
\begin{equation}
    \hat{S} = \theta_S([\mathcal{W}, \tilde{\mathcal{V}}^\Psi]).
\end{equation}
At a given 3D sampling rate, our point back-projection inference strategy adds a small but non-negligible cost relative to using only $\tilde{\mathcal{V}}^\Psi$: we must run an additional round of feature extraction and back-projection using $\Omega^f$, and the input dimension of $\theta_S$ must be doubled in order to receive the extra input features. We show in our experiments that this cost is tractable (see Section \ref{subsec:inference_time}).

We observe that the fine-grained features $\mathcal{W}$ can contain spurious high-frequency content, particularly near the borders of the 2D features where the CNN output is unreliable due to edge effects. To reduce artifacts, we down-weight $\{F_i^f\}$ near the borders prior to back-projection using a weight $w = \sigma(l \cdot (min(\frac{d}{m}, 1) \cdot 2 - 1))$, where $d$ is the distance to the nearest border in pixels; $m = 20$ is a margin distance; $l = 6$ controls the falloff rate; and $\sigma(x) = \frac{1}{1 + e^{-x}}$.

\vspace{-0.4cm}
\subsubsection{Output resolution \& occupancy filtering}
\label{subsec:occupancy}

Our model can be sampled at any point in $\mathbb{R}^3$, and we choose to sample it on a regular grid at test time in order to support meshing with marching cubes \cite{marchingcubes}. The resolution of this grid can be determined arbitrarily without re-training, and we experiment with several output grid resolutions (see Fig. \ref{fig:resolution}). Naively, the cost of our point back-projection inference strategy grows cubically with the sampling rate. At high resolutions, it thus becomes expensive due to the cost of running the additional back-projection and $\theta_S$ densely over the full volume. To mitigate this, we predict the per-voxel occupancy $\hat{O}$ with an additional MLP: $\hat{O} = \theta_{O}(\mathcal{V})$. Then at test time we sample $\hat{S}$ only within voxels that are predicted to be occupied. While the asymptotic complexity is still cubic, in practice this greatly reduces the cost due to the prevalence of empty space.

\begin{figure*}[t]
\begin{center}
    \begin{overpic}
    [width=\linewidth]
    {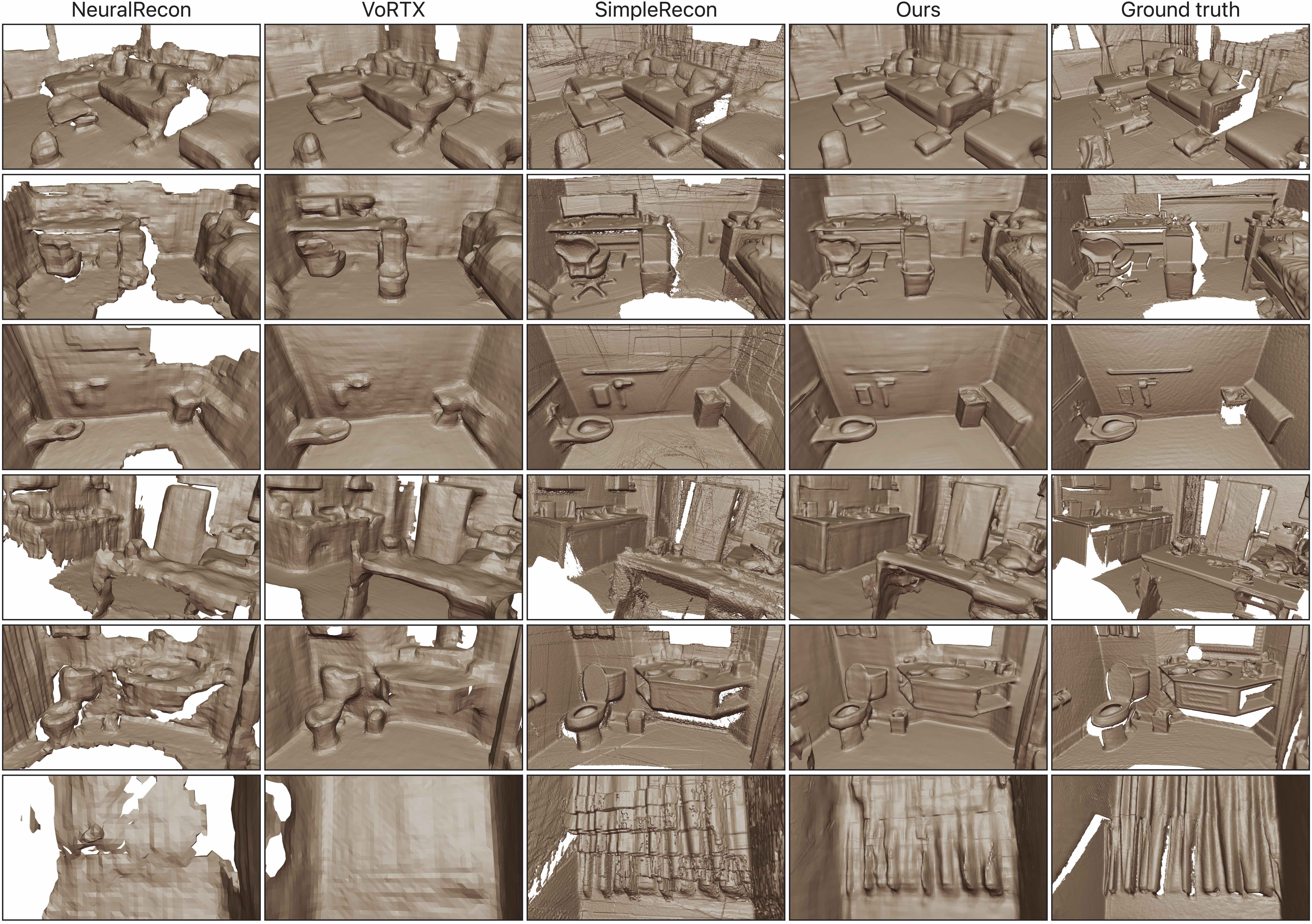}
    \end{overpic}
    \end{center}
    \vspace{-1em}
    \caption{
        \textbf{Qualitative results.} Comparison of our method with NeuralRecon \cite{neuralrecon}, VoRTX \cite{vortx}, and SimpleRecon \cite{simplerecon}. NeuralRecon and VoRTX capture consistent global structure but lose many details, whereas SimpleRecon recovers details but fails to keep geometry consistent across views, leading to duplicate surfaces after TSDF fusion. Our method produces the most complete, consistent reconstruction while preserving details.
    }

            \vspace{-0.5em}
\label{fig:qualitative}
\end{figure*}

\subsection{Training}

At training time, we require a ground-truth TSDF $S$ to supervise $\hat{S}$. For training on real scans, we use TSDF fusion \cite{curless1996volumetric} to generate $S$ on a discrete grid of points $X$ with resolution $\delta$, $X = \{x \in [ i \cdot \delta , j \cdot \delta, k \cdot \delta ] \}$, assuming the existence of a set of ground-truth depth maps $\{D_i\}$ corresponding to $\{I_i\}$. While ground truth depth can be noisy when acquired by sensors such as structured-light infrared scanners, we minimize artifacts by 1) using a large number of views to generate the ground-truth, 2) using an appropriate TSDF truncation distance following previous work \cite{atlas, vortx, neuralrecon}, and 3) discarding depths beyond the range where the accuracy starts to visibly degrade.

\vspace{-0.5cm}
\paragraph{Loss function.}
\label{subsec:loss}

We define the TSDF loss $\mathcal{L_S}$ following SG-NN \cite{sgnn} as 
\begin{equation}
    \mathcal{L}_S = \frac{1}{|X|} \sum_{x \in X} |t(\hat{S}(x)) - t(S(x)|,
\end{equation}
where $t(x) = \text{sign}(x) \cdot \text{ln}(|x| + 1)$. We define occupancy loss $\mathcal{L}_O$ using the standard binary cross-entropy, abbreviated $BCE$:
\begin{equation}
    \mathcal{L}_O = \frac{1}{|X|}\sum_{x \in X} BCE(O(x), \hat{O}(x))
\end{equation}
Following VoRTX \cite{vortx} we compute the ground-truth occupancy as
\begin{equation}
    O(x) = \oplus(\tilde{O}(x)); ~~ \tilde{O}(x) = \begin{cases}
    1,& \text{if } |S(x)| < 1\\
    0,              & \text{otherwise}
\end{cases}
\end{equation}
where $\oplus$ represents morphological dilation with a $3\times3\times3$ structuring element. Note, this definition of occupancy describes space near a surface \cite{sgnn, vortx, neuralrecon}, rather than space interior to an object. Our training loss $\mathcal{L}$ is then defined as
\begin{equation}
    \mathcal{L} = \mathcal{L}_S + \mathcal{L}_O.
\end{equation}
\paragraph{CNN backbone architectures.} Our 2D CNN architecture is a feature pyramid network \cite{lin2017feature} using EfficientNetV2-S \cite{efficientnetv2} as a backbone. This structure is shared for both $\Omega^c$ and $\Omega^f$, but the weights are duplicated and then trained independently. Despite their identical architectures, these networks learn strikingly different features (see Supp.). Our 3D CNN $\Psi$ is inspired by U-Net \cite{3dunet, unet}, see Supp. for details.

\paragraph{Implementation details.}

As the depth predictor $M$ we use our re-implementation of the MVS network SimpleRecon \cite{simplerecon}. We use a voxel size of 4cm in our model, and we train on volumetric scene chunks of size (3.84m$\times$3.84m$\times$2.24m). During training we select views by uniform random sampling over all views that at least partially observe the training chunk. At test time we fuse only keyframes, using the view selection strategy from DeepVideoMVS \cite{deepvideomvs}. Training takes 36 hours on eight Nvidia V100 GPUs.
\section{Experiments}
\label{sec:results}

\subsection{Dataset, baselines, and metrics}
\paragraph{Dataset.}

We validate our method by training and evaluating our model on the popular ScanNet dataset \cite{scannet}, which is composed of 1,613 indoor scans. We report all our metrics on the official test set containing 100 scans.
\vspace{-0.2cm}
\paragraph{Baselines.}

We compare our model with several previous works.
For end-to-end 3D reconstruction methods we select Atlas \cite{atlas}, NeuralRecon \cite{neuralrecon}, VoRTX \cite{vortx}, and TransformerFusion \cite{transformerfusion}.
We also compare to MVS depth-prediction, selecting SimpleRecon \cite{simplerecon}, which we re-implement.
In order to compare to SimpleRecon, we apply TSDF fusion \cite{curless1996volumetric} on the predicted depth maps to produce a 3D mesh.
\vspace{-0.2cm}
\paragraph{Metrics.}

We compute 3D metrics directly on the mesh reconstructions, and we compute depth metrics by rendering the meshes to generate predicted depth maps. Relative to prior works on depth estimation such as SimpleRecon \cite{simplerecon}, this is a slight change from the typical protocol of computing depth metrics directly on the raw depth maps. We believe our strategy is more indicative of performance in real-world applications that rely on the unified 3D reconstruction. Our definition for all metrics is the same as in Atlas~\cite{atlas} (see Supp. for details).

For computing 3D metrics we use the evaluation code from TransformerFusion~\cite{transformerfusion}, which includes a trimming protocol to avoid penalizing reconstructions for in-painting surfaces in unobserved regions. As noted in previous work \cite{transformerfusion, vortx}, precision or recall can easily be optimized individually at the expense of the other, as with accuracy and completeness. We thus emphasize Chamfer distance and F1 (i.e. F-score) as the most important 3D metrics, capturing this trade-off.

\begin{table}[t]
\begin{center}
\resizebox{\linewidth}{!}{
\begin{tabular}{l c c c c c c c}
    \toprule
    Method & Acc $\downarrow$ & Comp $\downarrow$ & Cham $\downarrow$ & Prec $\uparrow$ & Rec $\uparrow$ & F1 $\uparrow$ \\
    \midrule
Atlas \cite{atlas}
& 7.09 & 7.52  & 7.30 & 68.4 & 61.1 & 64.3
\\
NeuralRecon \cite{neuralrecon}
& 5.04 & 10.68 & 7.86 & 62.7 & 59.1 & 60.7
\\
3DVNet \cite{rich20213dvnet}
& 6.73 & 7.72  & 7.22 & 65.5 & 59.6 & 62.1 
\\
Transf. Fusion \cite{transformerfusion}
& {5.52} & 8.27  & 6.89 & {}{72.8} & 60.0 & 65.5 
\\
VoRTX \cite{vortx}
& \firstcell{4.32} & {7.52}  & \secondcell{5.92} & \secondcell{76.3} & {64.0} & \secondcell{69.5}
\\
SimpleRecon \cite{simplerecon}
& 6.43 & \secondcell{5.18}  & {5.80} & 66.0 & \secondcell{69.7} & {67.6}
\\
Ours
& \secondcell{5.25} & \firstcell{5.11}  & \firstcell{5.18} & \firstcell{78.0} & \firstcell{73.4} & \firstcell{75.5}
\\
\bottomrule
\end{tabular}
}
\end{center}
\caption{\textbf{3D reconstruction metrics for ScanNet.} We compare with recent work on mesh metrics defined in Atlas \cite{atlas}.  {\colorbox{red!35}{Best}} and \colorbox{orange!35}{Second}-best are highlighted. 
}
\label{tab:comparison}
\vspace{-0.2em}
\end{table}
\begin{figure*}[t]
    \begin{center}
        \begin{overpic}
            [width=0.9\linewidth]
            {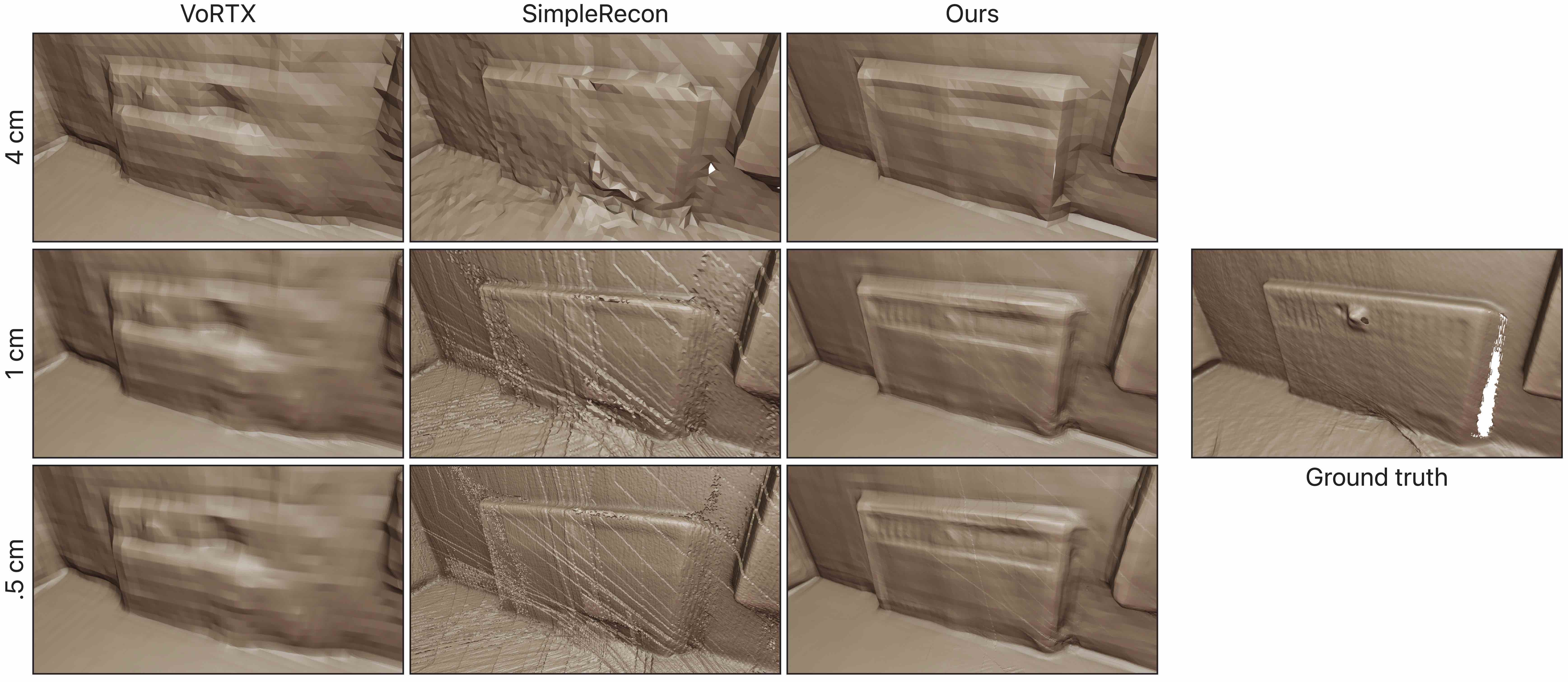}
        \end{overpic}
    \end{center}
    \vspace{-1em}
    \caption{
        \textbf{Qualitative results at various output resolutions.} With our method, we observe increased sub-voxel detail when increasing the output sampling rate from 4cm to 1cm, as well as the occasional appearance of small artifacts near view frustum boundaries. Beyond 1cm we observe no significant changes, suggesting we have reached our system's limit of detail given the 2D feature resolution. In contrast, upsampling the VoRTX outputs from 4cm slightly increases smoothness but adds no extra detail, and SimpleRecon shows the appearance of significant frustum boundary artifacts starting at 1cm.
    }
    \vspace{-0.5em}
    \label{fig:resolution}
\end{figure*}

\subsection{Results}

\paragraph{Qualitative Results.}
In Fig. \ref{fig:qualitative}, we show qualitative results of our method compared to baselines and ground truth. We observe that our method preserves more details while minimizing high-frequency artifacts. The surfaces in our reconstructions are smooth and consistent, gracefully unifying information from all views without the depth-disagreement artifacts or noticeable discontinuities visible in SimpleRecon.
VoRTX suffers from blurry reconstructions where fine details such as chair legs are completely lost, whereas our approach reconstructs these elements faithfully.

In Fig. \ref{fig:resolution} we show qualitative results at three output resolutions: 4cm, 1cm and 0.5cm. VoRTX is not tractable at high resolutions, so we upsample VoRTX's 4cm TSDF with linear interpolation, which does not increase the effective level of detail. For SimpleRecon, we can generate reconstructions at any desired resolution by decreasing the voxel size during TSDF fusion, and we note that artifacts and noise increase greatly at high resolution. In contrast, our results consistently preserve a high level of detail while avoiding noisy surfaces.

\begin{table}[t]
\begin{center}
\resizebox{\linewidth}{!}{
\begin{tabular}{l c c c c c c c}
    \toprule
    Method & L1 $\downarrow$ & AbsRel $\downarrow$ & SqRel $\downarrow$ & $\delta_{1.05} \uparrow$ & $\delta_{1.25} \uparrow$ & Comp. $\uparrow$\\
    \midrule
Atlas \cite{atlas} & 11.96 & 6.26 & 4.22 & 75.2 & 93.8 & \firstcell{98.8} \\
NeuralRecon \cite{neuralrecon} & 9.84  & 6.54 & 3.79 & 75.0 & 94.6 & 90.8 \\
VoRTX \cite{vortx} & {9.32}  & {5.92} & {3.65} & {79.0} & {94.9} & 96.1 \\
SimpleRecon \cite{simplerecon} & \secondcell{8.23}  & \secondcell{4.83} & \secondcell{2.66} & \secondcell{81.0} & \secondcell{96.8} & \secondcell{97.4} \\
Ours & \firstcell{6.91}  & \firstcell{4.24} & \firstcell{2.57} & \firstcell{86.6} & \firstcell{97.1} & {97.2} \\
\bottomrule
\end{tabular}
}
\end{center}
\caption{\textbf{2D metrics for ScanNet.} The 2D depth is rendered and metrics are computed in the same way as with Atlas \cite{atlas}.  We highlight the {\colorbox{red!35}{Best}} and \colorbox{orange!35}{Second} best with colors respectively.
}
\label{tab:main_2d}
\vspace{-1em}
\end{table}
\paragraph{Quantitative Results.}

Table \ref{tab:comparison} shows our 3D reconstruction metrics on the ScanNet dataset. We report metrics for our method and SimpleRecon at the relatively high resolution of 1cm because it is tractable to do so. We report metrics for the other baselines at their native resolution of 4cm, because increasing this resolution would add significant compute cost and engineering effort. For fairness, metrics for our method at 4cm are shown in Table \ref{tab:resolution}, and we note that the quantitative differences from 1cm are negligible. Our method achieves the best result in most metrics. In particular, it achieves the lowest Chamfer distance and highest F-score by large margins.

\newcolumntype{a}{>{\columncolor{brown!35}}c}

\begin{table}[t]
    \begin{center}
    \begin{footnotesize}
    \begin{tabular}{cccccc}
        \toprule
        & & \multicolumn{2}{c}{3D metrics} & \multicolumn{2}{c}{2D metrics} \\
        Method & Resolution & Cham$\downarrow$       & F1 $\uparrow$       & L1$\downarrow$ & $\delta_{1.05}\uparrow$ \\
        \midrule[0.1pt]
        \multirow{3}{*}{VoRTX~\cite{vortx}}
        & 4 cm & 5.92 & 69.5 & 9.32 & 79.0 \\
        & 1 cm & 5.91 & 69.6 & 9.27 & 79.0 \\
        & .5 cm & 5.91 & 69.5 & 9.29 & 79.0 \\
        \midrule[0.1pt]
        \multirow{3}{*}{SimpleRecon~\cite{simplerecon}}
        & 4 cm        & 5.51           & 68.6          & 8.40       & 80.4              \\
        & 1 cm       & 5.80           & 67.6          & 8.23       & 81.0              \\
        & .5 cm     & 5.94           & 66.9          & 8.67       &  80.7            \\
        \midrule[0.1pt]
        \multirow{3}{*}{Ours}
        & 4 cm        & 5.15           & 75.6          & 7.11       & 86.2              \\
        & 1 cm        & 5.18           & 75.5          & 6.91       & 86.6              \\
        & .5 cm      & 5.19           & 75.4          & 6.70       & 84.4             \\
        \bottomrule
    \end{tabular}
    \end{footnotesize}
    \end{center}
\caption{\textbf{Reconstruction metrics as a function of output resolution.} The 2D and 3D metrics show very little sensitivity to the output resolution, despite clear visual differences (see Figure \ref{fig:resolution}). This suggests that future work may require additional metrics to distinguish among high-quality reconstructions.
}
\label{tab:resolution}
\end{table}

\subsection{Ablation studies}
\label{subsec:ablation}

\begin{table}[t]
\begin{center}
\begin{footnotesize}
\begin{tabular}{c c c c c c c c}
    \toprule
& &  &  & \multicolumn{2}{c}{3D metrics} & \multicolumn{2}{c}{2D metrics} \\
& DG & PB & RTS 
& Cham $\downarrow$ & F1 $\uparrow$ & L1 $\downarrow$ & $\delta_{1.05}$ $\uparrow$  \\
\midrule[0.1pt]
(i) & \emptysquare & \emptysquare & \emptysquare
& 6.40 & 71.0 & 9.38 & 81.7 \\
(ii) & \emptysquare & \emptysquare & \checkedsquare
& 5.58 & 73.8 & 7.63 & 84.5 \\
(iii) & \emptysquare & \checkedsquare & \checkedsquare
& 5.80 & 72.4 & 8.26 & 83.0 \\
(iv) & \checkedsquare & \emptysquare & \checkedsquare
& 5.25 & 75.1 & 7.20 & 86.1 \\
(v) & \checkedsquare & \checkedsquare & \checkedsquare
& \textbf{5.18} & \textbf{75.5} & \textbf{6.91} & \textbf{86.6} \\
\bottomrule
\end{tabular}
\end{footnotesize}
\end{center}
\caption{\textbf{Ablation study.} We examine the quantitative improvement from the novel components of our method -- depth guidance (DG), point pack-projection (PB) and resolution-agnostic TSDF supervision (RTS). Relative to baseline (i), the largest gains come from RTS (ii). PB is helpful with DG enabled (v) but hurts metrics in the absence of DG (iii).
}
\label{tab:ablation_main}
\end{table}

In Table~\ref{tab:ablation_main} we compare ablations of the main novel components of our method. We note that resolution-agnostic TSDF supervision (RTS) and depth guidance (DG) both result in significant improvement across all metrics. Interestingly, point back-projection (PB) improves all metrics when DG is used ((v) vs. (iv)) but degrades them in the absence of DG ((iii) vs. (ii)). We interpret this as follows: the high-frequency content recovered by PB is locally accurate, but if the coarse alignment relative to ground truth is incorrect, then the added details actually reduce overall accuracy. DG helps to correctly localize the large structures, interacting constructively with PB to achieve the best performance.

\vspace{-0.4cm}
\paragraph{Depth guidance strategies}

\begin{table}[t]
\begin{center}
\resizebox{\linewidth}{!}{
\begin{tabular}{c l c c c c}
\toprule
 & & \multicolumn{2}{c}{3D metrics} & \multicolumn{2}{c}{2D metrics} \\
& Method & Cham $\downarrow$ & F1 $\uparrow$ & L1 $\downarrow$ & $\delta_{1.05}$ $\uparrow$  \\
\midrule[0.1pt]
(a) & Ours - TSDF (main)
& \textbf{5.18} & \textbf{75.5} & \textbf{6.91} & \textbf{86.6} \\
(b) & Ours - Density volume \cite{choe2021volumefusion} 
& 5.29 & 74.3 & 7.31 & 85.7 \\
(c) & Ours - Gaussian weight
& 5.47 & 73.2 & 7.48 & 85.1 \\
(d) & Ours - TSDF \& gaussian weight
& 5.52 & 73.8 & 7.24 & 85.8 \\
(e) & Ours - No depth guidance
& 5.80 & 72.4 & 8.26 & 83.0 \\
(f) & Ours - TSDF (no image features)
& 5.66 & 71.9 & 7.74 & 83.8 \\
\bottomrule
\end{tabular}
}
\end{center}
\caption{\textbf{Ablation of depth guidance strategies.} We compare our depth guidance strategy (a) to other forms of depth guidance: (b) a density fusion volume similar to VolumeFusion \cite{choe2021volumefusion}; (c) weighting the features using a Gaussian window centered on the predicted depth along each camera ray;  (d) combination of (b) and (c); (e) no depth guidance; and (f) TSDF depth guidance with no image features. We find that our TSDF guidance (a) achieves the best performance on all metrics.
}
\label{tab:ablation_depth}
\end{table}

In Table~\ref{tab:ablation_depth} we compare different ways to leverage the predicted depth maps to improve our reconstruction.
In row (b) we use the density fusion strategy introduced by VolumeFusion \cite{choe2021volumefusion}, and we observe that it does not perform as well as our method using TSDF fusion. We hypothesize that this is because the density does not encode free space information or inward vs. outward surface orientation.
In row (c) we test using the depth to directly modulate the image feature projection, using a Gaussian window to down-weight the image features far from the estimated depth.
Our experiments show that this manual weighting is an improvement relative to using no depth guidance at all, but that our TSDF fusion guidance yields the best results overall.
To explore the relative importance of the image features and the depth guidance, we ablate each one individually in rows (e) and (f), noting worse results in each case.

\subsection{Inference time}
\label{subsec:inference_time}

\OURS reconstructs the ScanNet test scenes at 4cm resolution in an average of 18s per scene using an Nvidia V100 GPU. This is composed of a per-frame time of 87ms for 2D feature extraction, depth estimation, and back-projection, plus a one-time TSDF extraction time of 1.1s including running the 3D CNN and output layers. Because we use a fixed voxel size, we can increase the output sampling rate with no increase to the per-frame time or 3D CNN time. As we apply higher spatial sampling rates, the cost of high-resolution inference increases proportionally to the number of occupied voxels due to back-projection and MLP execution at each sample point. Our average full-scene reconstruction time is 18s at 4cm, 21s at 2cm, 43s at 1cm. At the limit we test 0.5cm resolution which takes 3.6 minutes per scene on average. For faster inference times, the point back-projection can be disabled, resulting in an average time of 17s per scene at 1cm or 20s per scene at 0.5cm.
\section{Conclusion}
\label{sec:conclusion}

We have presented an end-to-end network producing detailed 3D reconstructions from posed images. We have demonstrated that our novel supervision (RTS) is key in enabling the network to learn fine details. We have also introduced large improvements by using a depth-prediction network to guide the back-projection (DG). Lastly, we have developed a novel architecture (PB) to allow the free selection of output resolution at test time without requiring additional training or 3D convolution levels. PB opens interesting future avenues in which the TSDF is not predicted on a regular grid, such as adaptive sampling rates or alternative polygonization algorithms beyond marching cubes.

\vspace{-0.4cm}
\paragraph{Limitations.}
While \OURS produces more accurate geometric details than prior work, limitations remain. 
Despite an improvement over the state-of-the-art, our approach still misses certain local fine structures. 
One reason is the limitation of the training data, in a large capture setting with pose noise and imperfect depth sensors. Another is that the forward-inference setting does not guarantee consistency with the input observations: to close this gap, future work may explore the hybridization of the techniques presented here with iterative optimization and differentiable rendering. Finally, despite our favorable inference times, running a 3D CNN over a dense feature volume is costly in terms of power and memory, which may be prohibitive for certain applications (sparse convolution may help, but it adds to implementation complexity and is not implemented on all platforms); increased efficiency is thus an important future goal.

{
    \small
    \bibliographystyle{ieee_fullname}
    \bibliography{references}
}

\end{document}